\documentclass{article} 
\usepackage{iclr2025_conference,times}

\usepackage{amsmath,amsfonts,bm}

\def\eqref#1{equation~\ref{#1}}

\def\1{\bm{1}}

\DeclareMathAlphabet{\mathsfit}{\encodingdefault}{\sfdefault}{m}{sl}
\SetMathAlphabet{\mathsfit}{bold}{\encodingdefault}{\sfdefault}{bx}{n}

\usepackage{xcolor}  
\definecolor{mydarkblue}{rgb}{0.68, 0.85, 1.0}
\definecolor{citecolor}{HTML}{0071bc}
\usepackage[pagebackref=false,breaklinks=true,colorlinks,citecolor=citecolor,bookmarks=false]{hyperref}
\usepackage{url}

\usepackage{url}            
\usepackage{booktabs}       
\usepackage{amsfonts}       
\usepackage{nicefrac}       
\usepackage{microtype}      

\usepackage{sidecap}
\usepackage{float}
\usepackage{graphicx}

\usepackage{afterpage}
\usepackage{multirow}
\usepackage{mathrsfs}

\usepackage{amsmath}
\usepackage{amssymb}
\usepackage{mathtools}
\usepackage[mathscr]{eucal}%
\usepackage{bm}
\usepackage{bbm}
\usepackage{relsize}
\usepackage{graphics}
\usepackage{wrapfig}
\usepackage{multirow}
\usepackage{enumitem}
\usepackage{tablefootnote}
\usepackage{pifont} 
\usepackage{subcaption}
\usepackage{array}
\usepackage{color, colortbl}
\definecolor{Gray}{gray}{0.9}
\definecolor{lightskyblue}{rgb}{0.53, 0.81, 0.98}
\definecolor{softblue}{rgb}{0.85, 0.91, 0.98}
\definecolor{mygray}{gray}{0.85}
\definecolor{lightblue}{RGB}{27,161,226}
\definecolor{bg_lightblue}{RGB}{218,232,252}
\usepackage{algorithm}
\usepackage{algorithmic}
\usepackage{pgf}
\usepackage{xinttools}
\usepackage{tikz}
\usetikzlibrary{tikzmark}
\usetikzlibrary{arrows.meta}
\usepackage{textcomp}
\usetikzlibrary{calc,shapes,arrows,positioning,automata,trees}
\usepackage{placeins} 
\usepackage{tabularx}
\usepackage{graphicx}
\usepackage{listings}
\def\ie{\emph{i.e.}}
\def\eg{\emph{e.g.}}

\def\etal{{\em et al.}}

\usepackage[font=small,labelfont=bf,tableposition=top]{caption}

\definecolor{Gray}{gray}{0.9}
\definecolor{lightGray}{gray}{0.95}
\definecolor{lightskyblue}{rgb}{0.53, 0.81, 0.98}
\definecolor{lightblue}{rgb}{0.68, 0.85, 0.9}
\definecolor{softblue}{rgb}{0.85, 0.91, 0.98}
\definecolor{mygray}{gray}{0.85}
\DeclareCaptionLabelFormat{andtable}{#1~#2  \&  \tablename~\thetable}

\usepackage{blindtext}

\newlength\savewidth

\makeatletter\renewcommand\paragraph{\@startsection{paragraph}{4}{\z@}
  {.5em \@plus1ex \@minus.2ex}{-.5em}{\normalfont\normalsize\bfseries}}\makeatother

\usepackage{bigdelim}
\usepackage{colortbl} 
\definecolor{mColor1}{rgb}{0.95,0.95,0.95}

\newcommand{\Ib}{{\boldsymbol I}}

\usepackage{float}

\definecolor{tabhighlight}{HTML}{e5e5e5}

\newcommand{\tableCellHeight}{1}
\newcommand{\tabstyle}[1]{
  \setlength{\tabcolsep}{#1}
  \renewcommand{\arraystretch}{\tableCellHeight}
  \centering
  \small
}
\usepackage{microtype}
\usepackage{graphicx}
\usepackage{booktabs} 

\usepackage[accsupp]{axessibility}  

\title{Prompt Diffusion Robustifies Any-Modality Prompt Learning}

\iclrfinalcopy

\author{
Yingjun Du\textsuperscript{1} \thanks{Work done during an internship at Cisco}, \hspace{5pt} 
Gaowen Liu\textsuperscript{2}, \hspace{5pt} 
Yuzhang Shang\textsuperscript{3}, \hspace{5pt} 
Yuguang Yao \textsuperscript{4}, \hspace{5pt} \\
\hspace{1pt}  \textbf{Ramana Kompella} \textsuperscript{\textbf{2}}, \hspace{5pt} 
\textbf{Cees G. M. Snoek}\textsuperscript{\textbf{1}} \\
\textsuperscript{1}University of Amsterdam  
\textsuperscript{2} Cisco Research \\
\textsuperscript{3}Illinois Institute of Technology
\textsuperscript{4} Michigan State University\\
}

\begin{document}

\maketitle

\begin{abstract}
Foundation models enable prompt-based classifiers for zero-shot and few-shot learning. Nonetheless, the conventional method of employing fixed prompts suffers from distributional shifts that negatively impact generalizability to unseen samples. This paper introduces \emph{prompt diffusion}, which uses a diffusion model to gradually refine the prompts to obtain a customized prompt for each sample. 
Specifically, we first optimize a collection of prompts to obtain over-fitted prompts per sample. Then, we propose a prompt diffusion model within the prompt space, enabling the training of a generative transition process from a random prompt to its overfitted prompt. As we cannot access the label of a test image during inference, our model gradually generates customized prompts solely from random prompts using our trained, prompt diffusion. Our prompt diffusion is generic, ﬂexible, and modality-agnostic, making it a simple plug-and-play module seamlessly embedded into existing prompt learning methods for textual, visual, or multi-modal prompt learning.
Our diffusion model uses a fast ODE-based sampling strategy to optimize test sample prompts in just five steps, offering a good trade-off between performance improvement and computational efficiency.
For all prompt learning methods tested, adding prompt diffusion yields more robust results for base-to-new generalization, cross-dataset generalization, and domain generalization in classification tasks tested over 15 diverse datasets.
\end{abstract}

\section{Introduction}
\label{sec:intro}

Foundation models trained on a diverse set of image-text pairs that encapsulate a virtually limitless vocabulary of real-world concepts~\citep{radford2021learning, jia2021scaling, li2022fine}, have demonstrated remarkable adaptability across various downstream tasks~\citep{lin2014microsoft, li2022end, li2023winner, zhang2022magic, zhang2024vision}. These models perform zero-shot image classification by filling in a predefined prompt template (\eg, “\texttt{a photo of a [CLASS]}”) with specific class names for the text encoder.
Despite their effectiveness in generalizing to new tasks, performance can be affected by minor alterations in the wording of prompt templates,~\citep{COOP}. 
Rather than manually creating hand-made prompts, several new prompt learning techniques in natural language processing \citep{lester2021power,liu2021gpt} and computer vision \citep{COOP, cocoop, jia2022visual, khattak2023maple, CoPrompt, li2024promptkd} have been suggested, which focus on learning a set of soft prompts with the aid of a small amount of labeled data. However, training a model with such deterministic prompts often results in overfitting, causing the model to focus too much on the training data, which affects its ability to generalize. These methods usually fail when a considerable distribution shift between training and test data leads to suboptimal generalization performance. 
We propose generating a distribution of prompts for each sample, employing a probabilistic approach that effectively incorporates visual (domain) information in a manner capable of learning and adaptation.

We are inspired by diffusion models~\citep{song2020denoising, zhou2024fast} that have emerged as a powerful
generative technique with broad applicability for tasks as diverse as image generation \citep{ho2020denoising}, video processing \citep{ho2022imagen}, and text generation \citep{gong2022diffuseq}. The core principle
behind diffusion involves an iterative refinement of the data distributions, transitioning from
a simple initial distribution to the desired target distribution. This iterative improvement process transforms the simple initial distribution into a series of sub-transformations, making it a versatile tool suitable for various tasks. \begin{wrapfigure}{r}{0.6\linewidth}
    \centering
    \includegraphics[width=0.95\linewidth]{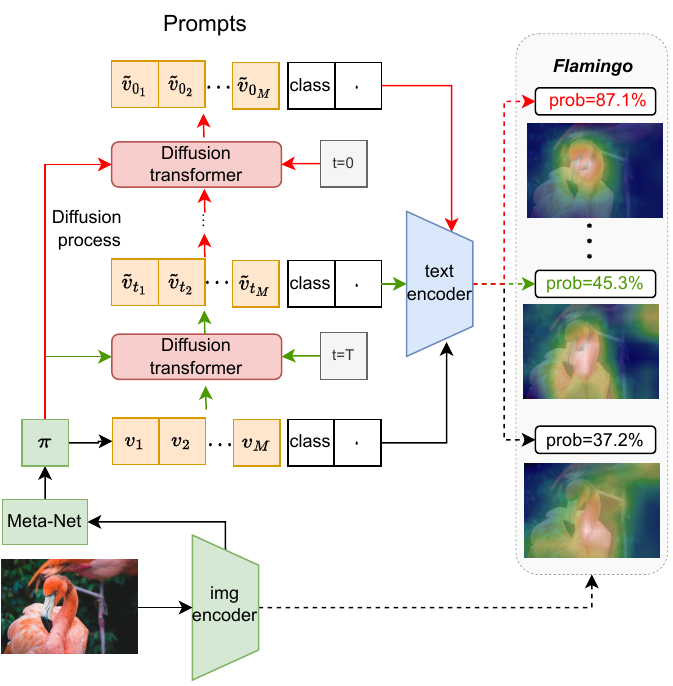}
\caption{
\textbf{Prompt diffusion} enhances traditional prompt learning methods such as CoCoOp~\citep{cocoop}  by introducing a diffusion process within the prompt space (colored arrows). Unlike deterministic prompt learning methods (black arrows), we employ a diffusion transformer to refine the prompts gradually. This process creates tailored prompts for each sample, complementing and augmenting existing prompting methods to achieve higher prediction accuracy through stronger generalization. }
    \label{fig:intro_diff}
\vspace{-5mm}
\end{wrapfigure}To the best of our knowledge, we are the first to introduce diffusion models into prompt learning. Our process of generating prompts through a diffusion model is depicted in Figure~\ref{fig:intro_diff}. Our prompt diffusion involves gradually refining the prompts with a diffusion transformer, which leads to the development of custom prompts tailored to each sample, thereby enhancing the accuracy of predictions and robustifying their generalization across downstream tasks.

In this paper, we make three contributions.
\textit{First}, we propose a prompt diffusion method based on the
transformer within the prompt space, enabling the learning of a generative pathway that seamlessly transitions from a random prompt to its personalized prompt.
Rather than relying on a single static prompt acquired for the entire dataset, our prompt diffusion can learn and evolve from noise to the example prompt throughout the training process. These personalized prompts are adept at generalizing the unique domain characteristics inherent in each sample, thus enhancing the model's ability to generalize. 
\textit{Second}, to better deploy prompt diffusion, we propose a per-sample overfitting strategy to obtain ``optima'' prompts for each data sample, allowing our diffusion transformer to effectively navigate the transition from general to highly personalized prompts within the training phase.
\textit{Third}, our prompt diffusion approach is versatile, adaptable, and modality-agnostic, which makes it easily integrated as a plug-and-play module within existing prompt learning techniques. This includes methods specialized for text-based prompts (\eg, CoCoOP~\citep{cocoop}), visual prompts (\eg, VPT~\citep{jia2022visual}), as well as three approaches that combine both text and visual inputs (\eg, MaPLe~\citep{khattak2023maple}, PromptSRC~\citep{khattak2023self}, and CoPrompt~\citep{CoPrompt}). 
Our diffusion model leverages a state-of-the-art fast ODE-based sampling strategy~\citep{zhou2024fast} that optimizes test sample prompts in just five steps, achieving an effective balance between performance enhancement and computational efficiency. To validate the effectiveness of our method, we conduct extensive testing across three common prompt learning experimental setups over 15 datasets: base-to-new generalization, cross-dataset generalization, and domain generalization. Adding prompt diffusion yields more robust results for all prompt learning methods tested.

\section{Related Work}


\noindent\textbf{Foundation models.} Foundation models developed through training on a wide and varied collection of image-text pairs, capture a nearly boundless array of concepts from the real world \citep{radford2021learning, jia2021scaling, li2022fine, schneider2024foundation, xu2024survey}, and have exhibited exceptional versatility in numerous downstream tasks \citep{lin2014microsoft, li2022end, li2023winner, zhang2022magic, zhang2024vision}. Foundation models can be  categorized into four types: 
1) Masked language modeling, as investigated in studies such as \citep{vilt,vilbert}, 2) Masked region prediction exemplified by \citep{lxmert,su2019vl}, 3) Image-text matching addressed by works like \citep{lxmert,vilt}, and 4) Contrastive learning, with notable references including \citep{CLIP,jia2021scaling,li2021align,wenlan}.
Numerous studies have demonstrated improved performance in tasks such as few-shot image recognition \citep{gao2021clip, zhang2021tip, kim2022how}, object detection \citep{li2024learning, Maaz2022Multimodal, zhou2022detecting, gu2021open, zang2022open, cheng2024yolo}, and segmentation \citep{li2024omg, rao2022denseclip, li2024transformer, luddecke2022image} using tailored methods. In this paper, we introduce a novel plugin designed to unify different prompt-learning approaches
to address the issues of prompt engineering in traditional foundation models, aimed at solving the base-to-new, cross-dataset, and domain generalization of visual recognition problems.

\noindent\textbf{Prompt learning.} Prompt learning, originally introduced in the natural language processing community~\citep{shin2020autoprompt, jiang2020can, liu2023pre}, involves applying a fixed function to input tokens to provide task instructions to the model. In the computer vision community, prompt learning has been explored in various forms, including textual prompts~\citep{COOP, cocoop, derakhshani2023bayesian, lu2022prompt, zhu2023prompt}, visual prompts~\citep{jia2022visual, ge2022domain, wang2022learning, bahng2022exploring, li2024visual, yang2024fine}, and multi-modal prompts~\citep{khattak2023maple, lee2023multimodal, li2023efficient, CoPrompt, li2024promptkd}.
1) Textual prompt learning, as pioneered by CoOp~\citep{COOP} and CoCoOp~\citep{cocoop}, fine-tunes a CLIP vision-language model \citep{CLIP} for few-shot transfer by optimizing a continuous set of prompt vectors within its language branch.  Bayesian prompt learning~\citep{derakhshani2023bayesian} formulated prompt learning as a variational inference problem and demonstrated its ability to generalize unseen classes at the expense of base class accuracy.
2) Visual prompt tuning~\citep{jia2022visual} introduces task-specific learnable prompts in the input visual space while keeping the pre-trained backbone fixed, optimizing them using the downstream task's label.
3) Multi-modal prompt learning~\citep{khattak2023maple, khattak2023self, li2024promptkd, xiao2024any} applied prompt learning in both vision and language branches to improve the alignment between the vision and language representations.
In contrast to previous prompt learning methods, this paper introduces modality-agnostic prompt diffusion, which leverages a diffusion model to generate prompts gradually. Our method serves as a simple plug-and-play module that seamlessly integrates with existing prompt learning methods, whether textual, visual, or multi-modal.

\section{Preliminaries}

Before detailing our prompt diffusion, we first present the technical background on the CLIP foundation model, prompt-based learning, and diffusion models. 

\noindent\textbf{Contrastive Language-Image Pre-Training (CLIP).} The objective of CLIP~\citep{CLIP} is to train an image encoder $f_I$ and a text encoder $g_T$ through contrastive pre-training  using a large set of paired images and texts. This encourages the encoders to align corresponding image-text pairs in a shared semantic space. After pre-training, CLIP exhibits the capacity for zero-shot visual recognition by casting classification as an image-text matching task. Specifically, the term ``\texttt{[CLASS]}'' is utilized as a placeholder within a prompt template (\eg, ``\texttt{a photo of a [CLASS]}'') for the text encoder $g_T$. If we let $g_T(\mathbf{T}_i)$ represent text features extended for class $i$, the classification probability for class $i$ given an image $\mathbf{I}$ is:
\begin{equation}
    p(y {=} i | \mathbf{I}) {=} \frac{\exp(\langle g_T(\mathbf{T}_i), f_I(\mathbf{I}) \rangle / \tau)}{\sum_{j{=}1}^{K} \exp(\langle g_T(\mathbf{T}_j), f_I(\mathbf{I}) \rangle / \tau)},
    \label{eq:predictions}
\end{equation} 
where $\langle g_T(\mathbf{T}_i), f_I(\mathbf{I}) \rangle$ denotes the cosine similarity between the image feature $f_I(\mathbf{I})$ and the class-specific text feature $g_T(\mathbf{T}_i)$ for the $i$-th class, $K$ the total number of classes, and $\tau$ the temperature parameter optimized during training.

\noindent\textbf{Prompt-based learning} enhances the transferability of the CLIP model by avoiding the need for prompt engineering. Instead, it enables automatic learning of prompts with a few samples from a downstream task. CoOp~\citep{COOP} introduces and refines a set of $M$ continuous context vectors $\bm{V}{{=}}\{\bm{v}_1, \bm{v}_2, \ldots, \bm{v}_M\}$ as the learnable prompt. The prompt $\bm{T}_i{{=}}\{\bm{v}_1, \bm{v}_2, \ldots, \bm{v}_M, \bm{c}_i\}$ is a concatenation of the learnable context vectors $\bm{V}$ and the class token embedding $\bm{c}_i$, which is then inputted to the text encoder $g_{T}(\cdot)$. CoOp tailors the static context vectors $\bm{V}$ by minimizing the negative log-likelihood for the correct class token:
        \begin{equation}
    \mathcal{L}_{\text{CE}}(\bm{V}){{=}}-\sum_i \bm{y}_i \log p(\bm{T}_i | \bm{I}),
    \label{eq:ce}
\end{equation}
Here, $\bm{y}_i$ denotes the one-hot ground-truth label for class $i$. In downstream tasks, the pre-trained model parameters remain frozen, allowing the learnable prompt vectors $\bm{V}$ to be efficiently optimized through the minimization of the cross-entropy loss with only a limited number of samples.

\noindent\textbf{Diffusion model.} 
In denoising diffusion probabilistic models~\citep{ho2020denoising}, $q(\mathbf{x}_t|\mathbf{x}_{t-1})$, is characterized as a Markov chain that progressively introduces Gaussian noise at each time step $t$, beginning with a clean image $\mathbf{x}_0 \sim q(\mathbf{x}_0)$. 
The \textit{forward} diffusion process is formulated as: 
   \begin{align}
     q(\mathbf{x}_{T}|\mathbf{x}_0):{=}\prod_{t{=}1}^{T}q(\mathbf{x}_t|\mathbf{x}_{t-1}),
\end{align} 
where $ q(\mathbf{x}_t|\mathbf{x}_{t-1}):{=}\mathcal{N}(\mathbf{x}_t;\sqrt{1-\beta_t}\mathbf{x}_{t-1},\beta_t\Ib)$, $\{\beta\}_{t{=}0}^{T}$ is a variance schedule.
By defining $\alpha_t {:{=}} 1 {-} \beta_t$ and $\bar{\alpha_t} {:{=}} \prod_{s{=}1}^t\alpha_s$, the forward diffused sample at time step $t$, denoted as $\bm x_t$, can be generated in a single step as $ \mathbf{x}_t{=}\sqrt{\bar{\alpha}_t}\mathbf{x}_0+\sqrt{1-\bar{\alpha}_t}\boldsymbol{\epsilon}$.

The \textit{reverse} process of the diffusion model learns to maximize the variational lower bound using parameterized Gaussian transitions, $p_\theta(\mathbf{x}_{t-1}|\mathbf{x}_t)$. Consequently, the reverse process is approximated as a Markov chain with the learned mean and fixed variance, starting from random noise $\mathbf{x}_T \sim  \mathcal{N}(\mathbf{x}_T; \textbf{0}, \boldsymbol{I})$.
The diffusion model is trained by optimizing the following objective function:
\begin{small}
    \begin{align}
    \label{eq:objective}
    \mathcal{L}_{\theta}{=}\mathbb{E}_{t,\mathbf{x}_0,\boldsymbol{\epsilon}}\Big{[}\|\boldsymbol{\epsilon}-\boldsymbol{\epsilon}_\theta(\sqrt{\bar{\alpha}_t}\mathbf{x}_0+\sqrt{1-\bar{\alpha}_t}\boldsymbol{\epsilon},t)\|^2\Big{]}.
\end{align}
\end{small}

In the \textit{sampling} phase of diffusion, to sample from $p_\theta(\mathbf{x}_{t-1}|\mathbf{x}_t)$, one can perform the following:
\begin{align}\label{eq:reverse}
    \mathbf{x}_{t-1} {=}\frac{1}{\sqrt{\alpha_t}}\Big{(}\mathbf{x}_t-\frac{1-\alpha_t}{\sqrt{1-\bar{\alpha}_t}}\boldsymbol{\epsilon}_\theta(\mathbf{x}_t,t)\Big{)}+\sigma_t\boldsymbol{\epsilon}.
\end{align}

Based on the geometric property that each sampling trajectory approximately resides within a two-dimensional subspace embedded in a high-dimensional space, \citet{zhou2024fast} introduce the Approximate MEan-Direction Solver (AMED-Solver), a single-step ODE solver that predicts the mean direction at each sampling step. By appropriately selecting $s_n$ and $c_n$, the AMED-Solver achieves an approximation given by:
\begin{equation}
    \label{eq:approx2}
    \mathbf{x}_{t_n} \approx \mathbf{x}_{t_{n+1}} + c_n (t_n - t_{n+1}) \mathbf{\epsilon}_{\theta}(\mathbf{x}_{s_n}, s_n).
\end{equation}
This formulation provides a single-step ODE solver, and the DPM-Solver-2~\citep{lu2022dpm} can be derived by setting $s_n = \sqrt{t_n t_{n+1}}$ and $c_n=1$. Unlike typical approaches that operate on images, our prompt diffusion model directly optimizes prompts. Given that prompt learning in vision-language tasks aims for faster and more accurate image classification, our proposed prompt diffusion, built upon the AMED-Solver, enables more rapid image classification during inference time.
\section{Methodology}
This section outlines our approach to training prompt learning via our proposed prompt diffusion model. We begin by explaining how to generate sample-specific overfitted prompts in Section~\ref{sec:overfit}. Next, we introduce prompt diffusion during both the training and testing phases to obtain diffused prompts in Section~\ref{sec:adopt}.

\subsection{Per-sample prompt overfitting }
\label{sec:overfit}

\begin{SCfigure}
    \centering\includegraphics[width=0.5\linewidth]{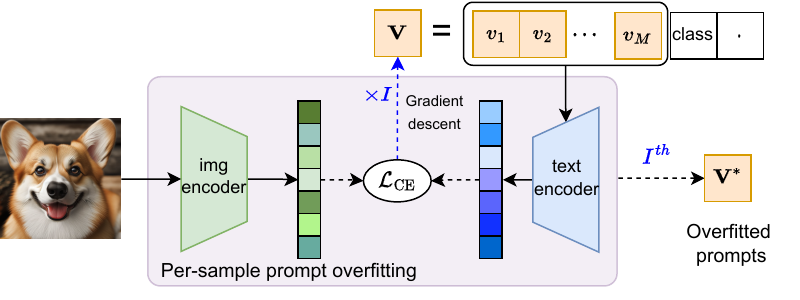}
\caption{\textbf{Per-sample prompt overfitting} for textual prompt learning. Through a minimal number of iterations \( I \) using gradient descent, we successfully derive overfitted prompts for each sample in the dataset. These overfitted prompts act as a ``ground truth'' for the prompts of each sample, enabling our proposed diffusion transformer to grasp the transition from generic prompts to highly personalized overfitted prompts.
}
    \label{fig:overfitting}
\end{SCfigure}

Our approach begins by fine-tuning various prompts to achieve individualized overfitting for each data sample. This ensures the precise generation of prompts that are tailored to specific instances. Specifically, when dealing with an image represented as $\bm{x}$, we aim to obtain a set of prompts, denoted as $\bm{V}^{*}$, which have been explicitly overfitted to that sample. We feed both the image $\bm{x}$ and the initial prompts $\bm{V} {=} \{v_1, v_2, \ldots, v_M\}$ into various prompt learning models and then employ iterative gradient descent on Eq.~(\ref{eq:ce}) to optimize the set of prompts, resulting in $\bm{V}^* {=} \{v_1^*, v_2^*, \ldots, v_M^*\}$. These optimized prompts can be considered as the ``optima'' prompts for each sample. Note that the intermediate loss is solely adjusted to achieve overfitted prompts in this process. Afterward, the gradient information for the learnable prompts will be discarded without optimization incorporated into the final loss.
We illustrate this per-sample prompt overfitting for textual prompt learning with CoCoOp~\citep{cocoop} in Figure~\ref{fig:overfitting}.

Once we obtain the overfitted prompts, our objective is to train the model using random prompts about these overfitted prompts. This is necessary because we cannot access the overfitted prompts during the testing stage. Therefore, in the following section, we use the diffusion model to learn the generative process of sample-specific prompts, thus robustifying the generalizability of the prompts for each sample.

\subsection{Prompt diffusion}
\label{sec:adopt}
We leverage the diffusion model~\citep{song2020denoising} to model textual, visual, or multi-modal prompts. In our implementation, we adapt the diffusion process to incrementally denoise and refine overfitted prompts, thereby enhancing the generative quality and coherence of the prompts.
Consequently, we introduce the concept of modality-agnostic prompt diffusion, a novel method that incrementally crafts sample-specific prompts for each instance. This methodical generation of prompts enhances their overall quality, ensuring that each prompt is optimally tuned to the nuances of its corresponding sample. This adaptive approach is designed to fine-tune the diffusion process, allowing for a more targeted and effective prompt generation that elevates the efficacy of the model's performance.

\noindent\textbf{Training phase.} During the initial training stage, we obtain the overfitted prompts $\bm{V}^*$ of individual samples via our proposed per-sample prompt overfitting. Then, the diffusion model is used to progressively approximate the overfitted prompts, from a Gaussian noise vector $\tilde{\bm{V}}_T \sim \mathcal{N}(0,\bm{I})$, which possesses the exact dimensions as $\bm{V}^*$.
The approximation process iterates through the noise vectors $\tilde{\bm{V}}_t^*$, with $t$ representing the diffusion step from $T$ to 0. This process leads to the reconstruction of \(\tilde{\bm{V}}_0\), which is expected to closely mirror the overfitted prompt associated with the particular sample being analyzed.

\begin{figure}[t]
    \centering\includegraphics[width=0.9\linewidth]{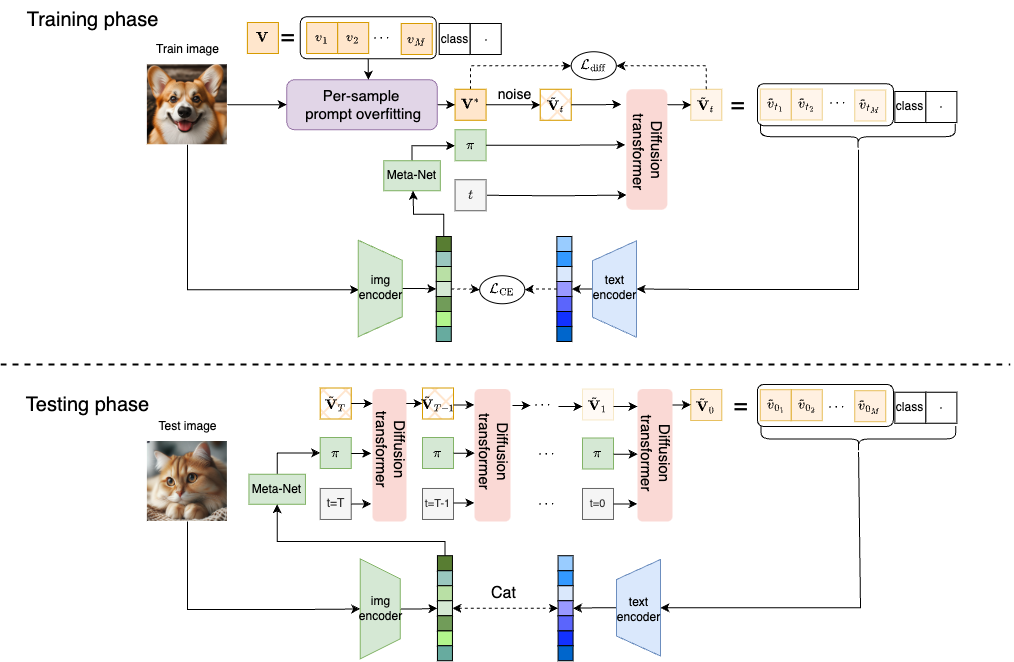}
\caption{\textbf{Prompt diffusion.} (1) Training by generating prompts that are initially overfitted using per-sample overfitting. These prompts are then subjected to a noise injection before entering the forward diffusion process. The inputs for diffusion include noisy prompts $\tilde{\bm{V}_t^*}$, the image features $\pi$, and a randomly chosen time step $t$, which leads to the generation of diffused prompts $\tilde{\bm{V}_t}$. After training, the diffusion transformer can convert generic prompts into their overfitted counterparts for each sample. (2) During testing,  the sampling process begins with an initial random noise $\tilde{\bm{V}}_{T}$, which is gradually refined into diffused prompts $\tilde{\bm{V}}_{t}$. At each time step $t$, the sampling process incorporates the previous state $\tilde{\bm{V}}_{t-1}$, test image features $\pi$, and current time step $t$ as inputs. The resulting diffused prompts $\tilde{\bm{V}_0}$ are then employed to make test sample predictions. Throughout $T$ with our diffusion transformer, the vanilla prompts are adapted into customized prompts that contain more specific information about the test sample, thereby enhancing prediction accuracy.}
    \label{fig:training_test}
    \vspace{-4mm}
\end{figure}

Specifically, throughout the forward diffusion phase at an increment in time $t$, we
derive the overfitted prompts $\bm{V}_t^*$.  
Subsequently, the noised prompts, denoted as $\tilde{\bm{V}}_t$, and the training image feature $\pi$ - extracted through a lightweight neural network, Meta-Net $\pi(\theta)$ \citep{cocoop} - are utilized to create a conditional token for each input and the temporal timestep $t$. These are then inputted into the diffusion transformer.
This process yields the interim diffused prompts $\tilde{\bm{V}}_t$. These prompts
then, the token $[\texttt{CLASS}]$ is synergized and integrated into the text encoder to generate the corresponding text features. The prediction of the final classification outcome for the training image is then conducted by Eq.~(\ref{eq:predictions}). For each sample, our diffusion model encapsulates a dual-component objective comprising the variational lower bound $\mathcal{L}_{\rm{diff}}$ for the diffusion model and the cross-entropy loss $\mathcal{L}_{\rm{CE}}$. The overarching schema of our training scheme is depicted at the top of Figure~\ref{fig:training_test}.

 The objective function, the simplified variational lower bound, aims to predict the denoised overfitted prompts accurately. Formally, the loss function is given by:
\begin{equation}
\mathcal{L}_{\text{diff}} = \left\lVert {\bm{V}}^* - \tilde{\bm{V}}_\theta\left(\sqrt{\bar{\alpha}_t}{\bm{V}}^* + \sqrt{1-\bar{\alpha}_t}\boldsymbol{\epsilon}, \pi, t\right) \right\rVert^2,
\label{eq:velbo}
\end{equation}
where $\tilde{\bm{V}}_\theta(\cdot, \cdot, \cdot)$ denotes the function parameterized by the transformer architecture~\citep{vaswani2017attention}. This function processes the input comprising the original overfitted prompts ${\bm{V}}^*$, image feature $\pi$, and the diffusion time step $t$. The efficacy of our model is measured by its ability to minimize this loss, thus accurately reconstructing the overfitted prompts from their noised counterparts.
By utilizing Eq.~(\ref{eq:ce}), we derive the final prediction $\hat{\mathbf{y}}$ using diffused prompts $\tilde{\bm{V}}_t$. The final objective is:
     \begin{equation}
\label{eq:final_loss}
\begin{aligned}
    \mathcal{L}_{\rm{final}} &= \sum_{(x, y)} \Big[ - \mathbb{E}{q_{(\tilde{\bm{V}}_t|\tilde{\bm{V}}_{t+1}, \pi)}}\big[\log p(\mathbf{y}|\mathbf{x},\tilde{\bm{V}}_t)\big] \\
&+ \beta \left\lVert {\bm{V}}^* - \tilde{\bm{V}}_\theta\left(\sqrt{\bar{\alpha}_t}{\bm{V}}^* + \sqrt{1-\bar{\alpha}_t}\boldsymbol{\epsilon}, \pi, t\right) \right\rVert^2,
\end{aligned}
\end{equation}  
where $\beta$ represents a hyperparameter.

In the supplemental materials, we provide the computational graph, which showcases the sequential steps of the forward and inverse diffusion processes on the prompts. Our method balances adaptability and informativeness by incorporating probabilistic prompts with the diffusion model.
Our model has also been applied to visual prompt tuning (VPT)~\citep{jia2022visual}  and multi-modal prompt learning (MaPLe, PromptSRC, and CoPrompt)~\citep{khattak2023maple, khattak2023self, CoPrompt}, generating visual prompts through a process identical to that used for generating text prompts.

\noindent\textbf{Testing phase.}
During the testing phase, the generation of overfitted prompts is infeasible due to the unavailability of test sample labels. Consequently, the diffusion sampling process begins with the introduction of Gaussian noise \( \tilde{\bm{V}}_T \) alongside the computed image feature set \( \pi \), followed by a systematic denoising procedure. To address an unseen test instance \( x \), initial image feature computations \( \pi \) are performed. After this, the noise vector \( \epsilon \) is drawn from a standard normal distribution \( \mathcal{N}(\textbf{0}, \textbf{I}) \) for each model and data/set. 
This ensures a diverse starting point for each prompt without using multiple models, training
multiple times, or employing different checkpoints.
These elements, comprising \( \tilde{\bm{V}}_T \), \( \pi \), and \( \epsilon \), are then supplied to the trained prompt diffusion model to derive intermediate diffused prompts \( \tilde{\bm{V}}_{T-1} \), represented by \( \tilde{\bm{V}}_{\theta}(\tilde{\bm{V}}_T, \pi, T) \). This 
iterative process unfolds over \( T \) steps, culminating in the acquisition of the terminal diffused prompts \( \tilde{\bm{V}}_0 {=} \tilde{\bm{V}}_{\theta}(\tilde{\bm{V}}_{1}, \pi, t_0) \). Upon retrieval of \( \tilde{\bm{V}}_0 \), integration with the text encoder occurs, facilitating the generation of relevant text features. The final stage involves the deployment of these features to predict the classification result for the test image, as delineated by Eq.~(\ref{eq:predictions}). The diffusion sampling framework throughout the testing phase is shown at the bottom of Figure~\ref{fig:training_test}.

\section{Experiments}\label{sec:exp}
We validate the effectiveness of our approach across three widely adopted scenarios for evaluating prompt learning in vision-language models: (1) base-to-new generalization, (2) cross-dataset generalization, and (3) domain generalization.

\subsection{Experimental setup}\label{sec:implementation}

\noindent\textbf{15 diverse datasets.} For  base-to-new generalization and cross-dataset gneralization, we follow CLIP~\citep{CLIP} and CoOp~\citep{COOP} to use 11 image classification datasets, \ie, ImageNet~\citep{imagenet} and Caltech101~\citep{caltech101} for generic object classification, OxfordPets~\citep{oxford_pets}, StanfordCars~\citep{stanford_cars}, Flowers102~\citep{flowers102}, Food101~\citep{food101} and FGVCAircraft~\citep{aircraft} for fine-grained image recognition, EuroSAT~\citep{eurosat} for  satellite image classification, UCF101~\citep{ucf101} for action classification, DTD~\citep{dtd} for texture classification, and SUN397~\citep{sun397} for scene recognition. 
For domain generalization, we 
follow CoOp~\citep{COOP} with ImageNet as the source dataset, and we select four variants of ImageNet: ImageNetV2~\citep{imagenetV2}, ImageNet-Sketch~\citep{imagenetSketch}, ImageNet-A~\citep{imagenetA} and ImageNet-R~\citep{imagenetR} as the target datasets.

\noindent\textbf{5 prompt learning baselines.} For comparative evaluation, we employ several established baselines:  (1) Textual prompt learning CoCoOp~\citep{cocoop}; (2) Visual prompt tuning (VPT)~\citep{jia2022visual}, representing the visual prompt learning method; (3) Multi-modal prompt learning (MaPLe~\citep{khattak2023maple}, PromptSRC~\citep{khattak2023self}), and CoPrompt~\citep{CoPrompt} employing prompt learning in both the visual and textual domains. Note that our method acts as a plugin that is easily integrated into each of these methods.

\begin{table}[t]
	\tabstyle{2.5pt}
 	\caption{\textbf{Base-to-new generalization.} Prompts are derived from base class examples. The harmonic mean (H) underscores the trade-off in generalization. Top-performing results are emphasized in {\color{blue}blue}. 
    By integrating our plugin within five different prompt learning methods, we consistently improve their average accuracy across 11 datasets, demonstrating enhanced performance over approaches without our plugin. }
	\vspace{-2mm}
 	\begin{subtable}[t]{.32\textwidth}
		\centering
		\caption{{{Average over 11 datasets}}.}
            \vspace{-2mm}
		    \scalebox{.58}{
		\begin{tabular}{ l cc|c}
			\toprule
			& Base & New & H \\
             \midrule
			VPT~\citep{jia2022visual}   &72.53  &72.34  &72.43  \\
               \rowcolor{lightGray}
			{+ {Prompt Diffusion}}   &{74.98}  &{74.97}  &{74.97}\\
			\midrule
                   CoCoOp~\citep{cocoop} &{80.47} & 71.69 &75.83 \\
                                \rowcolor{lightGray}
			{+ {Prompt Diffusion}}   &{81.35}  &{74.97}  &{78.02} \\
			\midrule
         MaPLe~\citep{khattak2023maple} & 82.28 & {75.14} & {78.55}  \\
                      \rowcolor{lightGray}
			{+ {Prompt Diffusion}}   &{{83.39}}  &{77.32}  &{80.24}   \\ 
   PromptSRC~\citep{khattak2023self} & 84.26 & {76.10} & {79.97} 
       
       \\
                      \rowcolor{lightGray}
			{+ {Prompt Diffusion}}   & {{85.74}}  & {78.97}  &{82.22}   \\ 
      CoPrompt~\citep{CoPrompt} & 84.00 & 77.23 &  80.48
       
       \\
                      \rowcolor{lightGray}
			{+ {Prompt Diffusion}}   &\color{blue}{{86.14}}  &\color{blue}{80.01}  &\color{blue}{82.96}   \\ 
			\bottomrule
		\end{tabular}}

	\end{subtable}
	\begin{subtable}[t]{.32\textwidth}
		\centering
		\caption{ImageNet.}
          \vspace{-2mm}
		    \scalebox{.58}{
		\begin{tabular}{l cc|c}
			\toprule
			& Base & New & H \\
			\midrule
			VPT~\citep{jia2022visual}   &74.45  &69.22  &71.74   \\
           \rowcolor{lightGray}
			{+ {Prompt Diffusion}}   &{74.97}  &{69.99}  &{72.39}   \\
			\midrule
   			CoCoOp~\citep{cocoop} & 75.98 &70.43 &73.10 \\
              \rowcolor{lightGray}
			{+ {Prompt Diffusion}}   &{76.46}  &{70.97} &{73.61}   \\
			\midrule
    MaPLe~\citep{khattak2023maple} & {76.66} & {70.54} & {73.47} \\
            \rowcolor{lightGray}
    			{+ {Prompt Diffusion}}   & {77.01}  &{71.03}  &{73.89}   \\   
                   PromptSRC~\citep{khattak2023self} &  77.60 & 70.73 &74.01 \\
                      \rowcolor{lightGray}
			{+ {Prompt Diffusion}}   &{{79.13}}  &{72.46}  &{75.65}   \\ 
         CoPrompt~\citep{CoPrompt} & 77.67 & 71.27 & 74.33
       
       \\
                      \rowcolor{lightGray}
			{+ {Prompt Diffusion}}   &\color{blue}{{80.73}}  &\color{blue}{73.25}  &\color{blue}{76.81}   \\ 
			\bottomrule
		\end{tabular}}
	\end{subtable}
	\begin{subtable}[t]{.32\textwidth}
		\centering
		\caption{Caltech101.}
          \vspace{-2mm}
		    \scalebox{.58}{
		\begin{tabular}{l cc|c}
			\toprule
			& Base & New & H \\
			\midrule 
			VPT~\citep{jia2022visual}   &96.92  &93.44  &95.15   \\
			 \rowcolor{lightGray} {+ {Prompt Diffusion}}  &{97.43}  &{94.23}  &{95.80}    \\
			\midrule
      			CoCoOp~\citep{cocoop} & 97.96 & 93.81 & 95.84 \\
			 \rowcolor{lightGray} {+ {Prompt Diffusion}}   &{98.12} &{94.97}  &{96.52}   \\
			\midrule
    MaPLe~\citep{khattak2023maple} & 97.74 & {94.36} & {96.02} \\
    			 \rowcolor{lightGray} {+ {Prompt Diffusion}}   &{97.25}  & {95.98}  &{96.61}   \\
                    PromptSRC~\citep{khattak2023self} &  98.10 &94.03 & 96.02  \\
                      \rowcolor{lightGray}
			{+ {Prompt Diffusion}}   & {98.08}  &\color{blue}{96.86}  &\color{blue}{97.47}   \\ 
         CoPrompt~\citep{CoPrompt} & 98.27 & 94.90 & 96.55 
       
       \\
                      \rowcolor{lightGray}
			{+ {Prompt Diffusion}}   &\color{blue}{{98.73}}  & {95.75}  & {97.22}   \\ 
			\bottomrule
		\end{tabular}}
	\end{subtable}
 	\begin{subtable}[t]{.32\textwidth}
		\centering
		\caption{{OxfordPets.}}
          \vspace{-2mm}
		    \scalebox{.58}{
		\begin{tabular}{l cc|c}
			\toprule
			& Base & New & H \\
\midrule
			VPT~\citep{jia2022visual}   &92.63  &94.96  &93.78   \\
			 \rowcolor{lightGray} {+ {Prompt Diffusion}}   &{93.17}  &{97.18}  &{95.14}    \\
			\midrule
      			CoCoOp~\citep{cocoop} &{95.20} & 97.69 &{96.43} \\
			 \rowcolor{lightGray} {+ {Prompt Diffusion}}   &{94.97}  &{97.98}  &{96.45}   \\
			\midrule
    MaPLe~\citep{khattak2023maple} & {95.43} & {97.76} & {96.58} \\
    			 \rowcolor{lightGray} {+ {Prompt Diffusion}}   & {95.96}  & {98.11}  & {97.02}   \\
                    PromptSRC~\citep{khattak2023self} & 95.33 & 97.30 & 96.30  \\
                      \rowcolor{lightGray}
			{+ {Prompt Diffusion}}   &{{95.44}}  &{98.05}  &{96.73}   \\ 
         CoPrompt~\citep{CoPrompt} & 95.67 & 98.10 & 96.87
       
       \\
                      \rowcolor{lightGray}
			{+ {Prompt Diffusion}}   &\color{blue}{{96.74}}  &\color{blue}{98.91}  &\color{blue}{97.81}   \\ 
			\bottomrule
		\end{tabular}}

	\end{subtable}
	\begin{subtable}[t]{.32\textwidth}
		\centering
		\caption{StanfordCars.}
          \vspace{-2mm}
		    \scalebox{.58}{
		\begin{tabular}{l cc|c}
			\toprule
			& Base & New & H \\
   \midrule
			VPT~\citep{jia2022visual}   &65.06  &74.68  &69.54   \\
			 \rowcolor{lightGray} {+ {Prompt Diffusion}}   &{65.75}  &{75.23}  &{70.17}   \\
			\midrule
      			CoCoOp~\citep{cocoop} & 70.49 & 73.59 & 72.01 \\
			 \rowcolor{lightGray} {+ {Prompt Diffusion}}   &{70.98}  & {75.32}  &{73.08}  \\
			\midrule
    MaPLe~\citep{khattak2023maple} & 72.94 & 74.00 & {73.47} \\
    			 \rowcolor{lightGray} {+ {Prompt Diffusion}}   & {73.11}  &{75.03}  &{74.06}   \\
                    PromptSRC~\citep{khattak2023self} &78.27 &74.97 & 76.58  \\
                      \rowcolor{lightGray}
			{+ {Prompt Diffusion}}   &\color{blue}{{80.14}}  &\color{blue}{76.15}  &\color{blue}{78.09}   \\ 
         CoPrompt~\citep{CoPrompt} & 76.97 & 74.40 & 75.66
       
       \\
                      \rowcolor{lightGray}
			{+ {Prompt Diffusion}}   & {{79.13}}  & {75.83}  & {77.44}   \\ 
			\bottomrule
		\end{tabular}}
	\end{subtable}
	\begin{subtable}[t]{.32\textwidth}
		\centering
		\caption{Flowers102.}
          \vspace{-2mm}
		    \scalebox{.58}{
		\begin{tabular}{l cc|c}
			\toprule
			& Base & New & H \\
			\midrule 
   			VPT~\citep{jia2022visual}   &76.23  &71.55  &73.82   \\
			 \rowcolor{lightGray} {+ {Prompt Diffusion}}   &{77.29}  &{72.33}  &{74.73}    \\
			\midrule
      			CoCoOp~\citep{cocoop} & 94.87 & 71.75 & 81.71 \\
			 \rowcolor{lightGray} {+ {Prompt Diffusion}}   &{94.17}  & {75.73}  &{83.95}   \\
			\midrule
    MaPLe~\citep{khattak2023maple} & 95.92 & 72.46 & {82.56} \\
    			 \rowcolor{lightGray} {+ {Prompt Diffusion}}   &{95.90}  &{73.14}  &{82.99}   \\
                    PromptSRC~\citep{khattak2023self} & 98.07 & 76.50 & 85.95  \\
                      \rowcolor{lightGray}
			{+ {Prompt Diffusion}}   &\color{blue}{{98.96}}  & {78.27}  & {87.41}   \\ 
         CoPrompt~\citep{CoPrompt} & 97.27 & 76.60 & 85.71
       
       \\
                      \rowcolor{lightGray}
			{+ {Prompt Diffusion}}   & {{98.73}}  &\color{blue}{78.49}  &\color{blue}{87.45}   \\ 
			\bottomrule
		\end{tabular}}
	\end{subtable}
  	\begin{subtable}[t]{.32\textwidth}
		\centering
		\caption{Food101.}
          \vspace{-2mm}
		    \scalebox{.58}{
		\begin{tabular}{l cc|c}
			\toprule
			& Base & New & H \\
			\midrule
			VPT~\citep{jia2022visual}   &89.27  &90.50  &89.88   \\
			 \rowcolor{lightGray} {+ {Prompt Diffusion}}   &{89.97}  &{92.12}  &{91.03}   \\
			\midrule
      			CoCoOp~\citep{cocoop} & 90.70 & 91.29 & 90.99 \\
			 \rowcolor{lightGray} {+ {Prompt Diffusion}}   &{90.21}  &{92.01}  &{91.10}   \\
			\midrule
    MaPLe~\citep{khattak2023maple} & {90.71} & {92.05} & {91.38} \\
    			 \rowcolor{lightGray} {+ {Prompt Diffusion}}   & {91.26}  &\color{blue}{93.11}  &\color{blue}{92.18}   \\
                    PromptSRC~\citep{khattak2023self} &  90.67 & 91.53 & 91.10  \\
                      \rowcolor{lightGray}
			{+ {Prompt Diffusion}}   & {90.74}  &{92.58}  &{91.65}   \\ 
         CoPrompt~\citep{CoPrompt} & 90.73 & 92.07 &91.40
       
       \\
                      \rowcolor{lightGray}
			{+ {Prompt Diffusion}}   &\color{blue}{{91.34}}  & {92.98}  & {91.25}   \\ 
			\bottomrule
		\end{tabular}}
	\end{subtable}
	\begin{subtable}[t]{.32\textwidth}
		\centering
		\caption{FGVCAircraft.}
          \vspace{-2mm}
		    \scalebox{.58}{
		\begin{tabular}{l cc|c}
			\toprule
			& Base & New & H \\
			\midrule
   			VPT~\citep{jia2022visual}   &28.23  &32.21  &30.09   \\
			 \rowcolor{lightGray} {+ {Prompt Diffusion}}   &{28.82}  &{35.07}  &{31.64}    \\
			\midrule
      			CoCoOp~\citep{cocoop} &{33.41} & 23.71 & 27.74 \\
			 \rowcolor{lightGray} {+ {Prompt Diffusion}}   &{34.21}  &{35.27}  &{34.73}   \\
			\midrule
    MaPLe~\citep{khattak2023maple} & 37.44 & 35.61 & {36.50} \\
    			 \rowcolor{lightGray} {+ {Prompt Diffusion}}   & {37.11}  & {36.15}  & {36.62}   \\
                    PromptSRC~\citep{khattak2023self} & 42.73 & 37.87 & 40.15  \\
                      \rowcolor{lightGray}
			{+ {Prompt Diffusion}}   &\color{blue}{{44.81}}  & {39.98}  &\color{blue}{42.26}   \\ 
         CoPrompt~\citep{CoPrompt} & 40.20 & 39.33 & 39.76
       
       \\
                      \rowcolor{lightGray}
			{+ {Prompt Diffusion}}   & {{42.35}}  &\color{blue}{41.27}  & {41.80}   \\ 
			\bottomrule
		\end{tabular}}
	\end{subtable}
	\begin{subtable}[t]{.32\textwidth}
		\centering
		\caption{SUN397.}
          \vspace{-2mm}
		    \scalebox{.58}{
		\begin{tabular}{l cc|c}
			\toprule
			& Base & New & H \\
			\midrule 
			VPT~\citep{jia2022visual}   &75.14  &76.89  &76.00   \\
			 \rowcolor{lightGray} {+ {Prompt Diffusion}}   &{75.74}  &{77.82}  &{76.77} \\
			\midrule
      			CoCoOp~\citep{cocoop} & 79.74 & 76.86 & 78.27\\
			 \rowcolor{lightGray} {+ {Prompt Diffusion}}   &{80.14}  &{77.53}  &{78.81}   \\
			\midrule
    MaPLe~\citep{khattak2023maple} & {80.82} & {78.70} & {79.75} \\
    			 \rowcolor{lightGray} {+ {Prompt Diffusion}}   &{81.03}  &{79.54}  &{80.28}   \\
                    PromptSRC~\citep{khattak2023self} & 82.67 & 78.47 & 80.52  \\
                      \rowcolor{lightGray}
			{+ {Prompt Diffusion}}   & {{84.15}}  & {80.27}  & {82.16}   \\ 
         CoPrompt~\citep{CoPrompt} & 82.63 & 80.03 & 81.31
       
       \\
                      \rowcolor{lightGray}
			{+ {Prompt Diffusion}}   &\color{blue}{{84.71}}  &\color{blue}{81.97}  &\color{blue}{83.32}   \\ 
			\bottomrule
		\end{tabular}}
	\end{subtable}
  	\begin{subtable}[t]{.32\textwidth}
		\centering
		\caption{DTD.}
          \vspace{-2mm}
		    \scalebox{.58}{
		\begin{tabular}{l cc|c}
			\toprule
			& Base & New & H \\
			\midrule
   			VPT~\citep{jia2022visual}   &56.71  &57.25  &56.98   \\
			 \rowcolor{lightGray} {+ {Prompt Diffusion}}  &{58.43}  &{58.13}  &{58.28}  \\
			\midrule
      			CoCoOp~\citep{cocoop} &{77.01} & 56.00 & 64.85 \\
			 \rowcolor{lightGray} {+ {Prompt Diffusion}}   &{73.43}  &{60.19}  &{66.15}   \\
			\midrule
    MaPLe~\citep{khattak2023maple} & {80.36} & 59.18 & {68.16} \\
    			 \rowcolor{lightGray} {+ {Prompt Diffusion}}   &{80.25}  &{59.94}  &{68.62}   \\
                    PromptSRC~\citep{khattak2023self} & 83.37 & 62.97 & 71.75  \\
                      \rowcolor{lightGray}
			{+ {Prompt Diffusion}}   &\color{blue}{{85.71}}  & {65.07}  & {73.98}   \\ 
         CoPrompt~\citep{CoPrompt} & 83.13 & 64.73 & 72.79
       
       \\
                      \rowcolor{lightGray}
			{+ {Prompt Diffusion}}   & {{85.14}}  &\color{blue}{65.96}  &\color{blue}{74.33}   \\ 
			\bottomrule
		\end{tabular}}
	\end{subtable}
	\begin{subtable}[t]{.32\textwidth}
		\centering
		\caption{EuroSAT.}
          \vspace{-2mm}
		    \scalebox{.58}{
		\begin{tabular}{l cc|c}
			\toprule
			& Base & New & H \\
			\midrule
   			VPT~\citep{jia2022visual}   &67.57  &59.69  &63.39   \\
			+ {Prompt Diffusion}  &{67.26}  &{69.01}  &{68.13}  \\
			\midrule
      			CoCoOp~\citep{cocoop} & {87.49} & 60.04 & 71.21 \\
			+ {Prompt Diffusion}  &{88.13}  &{70.22}  &{78.16}   \\
			\midrule
    MaPLe~\citep{khattak2023maple} & {94.07} & {73.23} & {82.35} \\
    			 \rowcolor{lightGray} {+ {Prompt Diffusion}}   & {94.76}  &{73.34}  &{82.69}   \\
                    PromptSRC~\citep{khattak2023self} & 92.90 & 73.90 & 82.32\\
                      \rowcolor{lightGray}
			{+ {Prompt Diffusion}}   & {{93.94}}  & {76.07}  & {84.07}   \\ 
         CoPrompt~\citep{CoPrompt} & 94.60 & 78.57 & 85.84 
       
       \\
                      \rowcolor{lightGray}
			{+ {Prompt Diffusion}}   &\color{blue}{{94.98}}  &\color{blue}{80.17}  &\color{blue}{86.95}   \\ 
			\bottomrule
		\end{tabular}}
	\end{subtable}
	\begin{subtable}[t]{.32\textwidth}
		\centering
		\caption{UCF101.}
          \vspace{-2mm}
		    \scalebox{.58}{
		\begin{tabular}{l cc|c}
			\toprule
			& Base & New & H \\
			\midrule 
   			VPT~\citep{jia2022visual}   &75.65  &75.31  &75.48   \\
			 \rowcolor{lightGray} {+ {Prompt Diffusion}}   &{76.31}  &{76.23}  &{76.27}  \\
			\midrule
   			CoCoOp~\citep{cocoop} & {82.33} & 73.45 & 77.64 \\
			 \rowcolor{lightGray} {+ {Prompt Diffusion}}   &{81.97}  &{77.03}  &{79.42}   \\
			\midrule
    MaPLe~\citep{khattak2023maple} & 83.00 & {78.66} & {80.77} \\
    			 \rowcolor{lightGray} {+ {Prompt Diffusion}}   &{82.86}  &{79.64}  &{81.22}   \\
                    PromptSRC~\citep{khattak2023self} & 87.10 & 78.80 &82.74 \\
                      \rowcolor{lightGray}
			{+ {Prompt Diffusion}}   &\color{blue}{{88.21}}  & {79.91}  & {83.86}   \\ 
         CoPrompt~\citep{CoPrompt} & 86.90 & 79.57 & 83.07
       
       \\
                      \rowcolor{lightGray}
			{+ {Prompt Diffusion}}   & {{88.14}}  &\color{blue}{80.28}  &\color{blue}{84.03}   \\ 
			\bottomrule
		\end{tabular}}
	\end{subtable}

  \label{tab:base2new}
  \vspace{-5mm}
\end{table}

\noindent\textbf{Training details.} To ensure a fair comparison, we utilize the CLIP-ViT-B/16 as the base pre-training model for CoCoOp~\citep{cocoop}, and  VPT~\citep{jia2022visual}, setting the prompt token count to 4. This configuration is based on recommendations in~\citep{cocoop}, indicating optimal performance with a more concise context length. For MaPLe~\citep{khattak2023maple}, PromptSRC~\citep{khattak2023self} and CoPrompt~\citep{CoPrompt}, the prompt depth $M$ is adjusted to 9, and we configure the language and vision prompt lengths at two tokens each. In the diffusion preprocessing stage, we adapt the strategy of positional token assignment~\citep{dosovitskiy2020image} to the input prompts $\tilde{\bm{V}}^*$ and the image features $\pi$. Furthermore, the diffusion time step $t$ is encoded as a series of individual tokens, adopting a frequency-based vector representation scheme~\citep{mildenhall2021nerf}. We set the diffusion time step $t$ as 100 for our experiments. Our transformer-based model architecture is the same as the GPT-2 framework~\citep{radford2019language}. This includes a 12-layer transformer, a linear transformation, and an attention mechanism with 16 heads. The batch size is 32 for all prompt-based models, except for CoCoOp, which is trained with a batch size of 4. Each model leverages a learning rate 0.0035 applied through the SGD optimizer on a single NVIDIA A100 GPU for execution. Code will be made available.

\noindent\textbf{Evaluation setting.}
Across all experiments, we benchmark the models' performance in a 16-shot setting, standardizing the number of training epochs to 50 for each baseline and dataset. The appendix presents a 4-shot experiment, compares outcomes across different epochs, and evaluates various parameter-efficient approaches. For consistency, all results from learning-based methods are computed as an average over three random seeds.  

\subsection{Comparative experiments}

\noindent\textbf{Base-to-new generalization.}\label{sec:base2new}
Table~\ref{tab:base2new}  shows that various prompting methods, when combined with our prompt diffusion approach, consistently surpass the average performance across all datasets.
Regarding base class accuracy averaged across 11 datasets, our approach advances VPT, CoCoOp, MaPle, PromptSRC, and CoPrompt by 2.54\%, 1.08\%, 1.11\%, 2.25\% and 2.48\%, respectively, showcasing that our approach strengthens the adaptation of existing methods. When it comes to recognizing new classes, our approach also shows good improvement, with gains of 2.60\% for VPT, 1.26\% for CoCoOp,  2.78\% for MaPle  1.81\% for PromptSRC, and 2.87\% for CoPrompt, emphasizing its effectiveness in dealing with unseen samples.
Regarding the harmonic mean, which considers both base and new classes, our method retains a superior few-shot generalization capacity across all datasets compared to baseline models. Notably, CoPrompt, with our prompt diffusion, consistently outperforms all other methods across most datasets, demonstrating the advantages of using both modalities in prompt learning.
Our prompt diffusion, applied to different prompt learning models, consistently improves the outcomes by generating more informative and precise prompts.
%

\begin{table}[t]
 \caption{\textbf{Cross-dataset generalization.} Accuracy (\%) evaluation for prompts learned from the source dataset. Our plugin consistently enhances existing prompt learning methods, whether textual, visual, or multi-modal.}
 \vspace{-3mm}
\centering
    \scalebox{0.57}{
	\begin{tabular}{l c ccccccccccc}
		\toprule
		& Source & \multicolumn{11}{c}{Target} \\ \cmidrule(lr){2-2} \cmidrule(lr){3-13}
		& \rotatebox{45}{ImageNet} & \rotatebox{45}{Caltech101} & \rotatebox{45}{OxfordPets} & \rotatebox{45}{StanfordCars} & \rotatebox{45}{Flowers102} & \rotatebox{45}{Food101} & \rotatebox{45}{Aircraft} & \rotatebox{45}{SUN397} & \rotatebox{45}{DTD} & \rotatebox{45}{EuroSAT} & \rotatebox{45}{UCF101} & \rotatebox{45}{Average} \\
		\midrule
		VPT~\citep{jia2022visual}   &68.92 &93.07  &89.44   &64.77   &67.79  &84.91  &23.72  &66.16  &45.02  &37.74  &67.00 &63.96  \\
		            \rowcolor{lightGray} + Prompt diffusion &{70.23}		&{94.71}	&{90.93}	&{65.53}	&{68.93}	&{85.71}	&{24.81}	&{66.98}	&{46.16}	&{39.67}	&{67.91}	&{65.11}\\
		\midrule
		CoCoOp~\citep{cocoop}  & 71.02 & 94.43 & 90.14 &65.32 &71.88 & 86.06 &22.94 &67.36 &45.73 &45.37 &68.21 &65.74 \\
		            \rowcolor{lightGray} + Prompt diffusion &\color{blue}{71.98}		&{95.07}	& {91.11}	&{66.73}	& {73.52}	&{87.18}	&{22.23}	&{68.25}	&{46.84}	&{47.13}	&{69.53}	&{66.76}\\
		\midrule
    	MaPLe~\citep{khattak2023maple} & 70.72 & 93.53 & {90.49} & {65.57} & {72.23} & {86.20} & {24.74} & 67.01 & {46.49} & {48.06} & {68.69} & {66.30} \\
		            \rowcolor{lightGray} + Prompt diffusion & {71.23} & 95.98 & {92.49} & {67.17} & {74.13} & {88.24} & {26.23} & {69.43} & {47.95} & {49.73} & {69.53} & {68.09} \\
                  	PromptSRC~\citep{khattak2023self} & 71.27 &93.60 & 90.25 &65.70 &70.25& 86.15 &23.90 &67.10 &46.87 &45.50 &68.75& 65.81 \\
		            \rowcolor{lightGray} + Prompt diffusion & {71.73} & 96.01 & {93.13} & {68.12} & {73.71} & {88.31} & {26.14} & {70.21} & {48.35} & {48.15} & {70.24} & {68.23} \\
                                	CoPrompt~\citep{CoPrompt} & 70.80 & 94.50  &90.73& 65.67& 72.30& 86.43 &24.00 &67.57 &47.07 &51.90 &69.73 &67.00\\
		            \rowcolor{lightGray} + Prompt diffusion & {71.46} & \color{blue}{96.12} & \color{blue}{93.94} & \color{blue}{68.81} & \color{blue}{74.98} & \color{blue}{88.11} & \color{blue}{26.31} & \color{blue}{71.73} & \color{blue}{49.15} & \color{blue}{54.41} & \color{blue}{71.14} & \color{blue}{69.47} \\

		\bottomrule
	\end{tabular}}
 \label{tab:cross-dataset}
 \vspace{-5mm}
\end{table}

\noindent\textbf{Cross-dataset generalization.}\label{sec:cross-dataset}
Our study assesses how well models can adapt prompt learning from one dataset and apply it effectively to different datasets for cross-dataset generalization. We test the zero-shot transfer capabilities of the models on a wide range of 10 datasets. As shown in Table~\ref{tab:cross-dataset}, our prompt diffusion substantially improves the average transfer performance of models like  VPT, CoCoOp, MaPLe, PromptSRC and CoPrompt, with respective increases of1.15\%, 1.02\%, 1.79\%,  2.43\%, and 2.47\%.
These results not only confirm the effectiveness of our method in enhancing cross-dataset generalization but also highlight its versatility across various prompt learning methods.

\noindent\textbf{Domain generalization.}\label{sec:domain_generalization}
The performance of various ImageNet 
variants, which have a domain shift compared with the source dataset, is evaluated.
Table~\ref{tab:dg} summarizes these findings, highlighting not only the improvement in performance across  VPT, CoCoOp, MaPLe, PromptSRC, and CoPrompt but also the maintenance of performance on the source dataset itself. 
Interestingly, using prompt diffusion with CoCoOp performs better than all multi-modal prompt learning methods on the ImageNet-A dataset. This could be due 
to CoCoOp's emphasis on textual prompt learning, which may be more suited to these types of datasets. ImageNet-A images frequently display anomalies or atypical features, whereas
visual prompts could highlight deceptive or overly intricate aspects, making classification less accurate. Thus, when addressing 
real-world datasets like ImageNet-A, 
it is advantageous to use 
textual prompting
with our prompt diffusion.
\begin{wraptable}{r}{9.5cm}
\vspace{-3mm}
 \caption{\textbf{Domain generalization.} Accuracy (\%) evaluation on target datasets using prompts learned from a source dataset. Our method delivers consistent, prompt learning improvements across all datasets.}
    \scalebox{0.7}{
	\begin{tabular}{l cccccc}
		\toprule
		& Source & \multicolumn{5}{c}{Target} \\ \cmidrule(lr){2-2} \cmidrule(lr){3-7}
		& ImageNet & -V2 & -S & -A & -R &Average\\
		\midrule
  				VPT~\citep{jia2022visual}   &  68.92   &61.84  &47.64  &46.50   &75.86    &57.96            \\
		            \rowcolor{lightGray} + Prompt diffusion  &{70.23}        &{62.97}           &{48.77}            &{47.25}  &{77.06} &{59.01} \\
		\midrule
  CoCoOp~\citep{cocoop}  & 71.02 & 64.07 & 48.75 &50.63 &76.18 &59.91\\
		            \rowcolor{lightGray} + Prompt diffusion & \color{blue}{71.98}        & {65.28}           &{50.11}            &\color{blue}{52.23}   &{77.50} &{61.25} \\
		\midrule
		{MaPLe}~\citep{khattak2023maple} & 70.72 & {64.07} &{49.15} & {50.90}  & {76.98}  & {60.83}  \\
  		            \rowcolor{lightGray} + Prompt diffusion  &{ 71.23}        &{65.49}           &{50.46}            &{52.18}  &{78.31} &{62.36} \\
                		{PromptSRC}~\citep{khattak2023self} & 71.27 & 64.35& 49.55& 50.90 &77.80 & 60.65 \\
  		            \rowcolor{lightGray} + Prompt diffusion  &{ 71.73}        &\color{blue}{66.33}           &\color{blue}{51.21}            &{52.02}  & {79.86} &\color{blue}{62.88} \\
                                		{CoPrompt}~\citep{CoPrompt} & 70.80 & 64.25 &49.43 &50.50 &77.51 &60.42 \\
  		            \rowcolor{lightGray} + Prompt diffusion  &{ 71.46}        & {66.01}           &{50.71}            &{51.75}  &\color{blue}{80.76} & {62.30} \\
		\bottomrule 
	\end{tabular}}
\label{tab:dg}
\end{wraptable}
Since such datasets often contain natural images, textual prompts can
leverage the semantic context effectively. On the other hand, in scenarios involving a clear distribution shift (\eg~\textit{sketch}, \textit{cartoon}), employing a multi-modality prompt with our prompt diffusion is more effective.
From the results of these experiments, our method fosters a level of adaptability that allows models to maintain their initial generalizability even after being fine-tuned to limited datasets.

\subsection{Ablation experiments}

\noindent\textbf{Benefit of the diffusion model.}
To confirm that the performance gain of our model can be attributed to the diffusion model, we first conducted the experiments using MLP and transformers as non-generative models, using overfitted prompts as supervision.  We also compared it with three widely
used generative
models: generative adversarial networks (GAN)~\citep{goodfellow2014generative}, variational auto-encoders (VAE)~\citep{kingma2013auto}, and normalizing 
flows~\citep{rezende2015variational}. 
First, we obtain the overfitted prompts with per-sample prompt overfitting.
In the case of
non-generative models, the process involves solely using image features
$\pi$, and then employing overfitted prompts, $\bm{V}^*$, for supervising the training of MLP and transformer models.
The input is extended for GANs, VAE and normalizing flows to include image
features $\pi$ and a variable $\epsilon$  sampled from a standard normal distribution $\mathcal{N}(0, I)$.
These inputs, along with the overfitted
\begin{wraptable}{r}{5.5cm}
\vspace{-4mm}
  \caption{\textbf{Benefit of diffusion model} in the base-to-new generalization.}
  \vspace{-3mm}
\centering
\scalebox{0.65}{
	\begin{tabular}{l cc|c}
			\toprule
			& Base & New & H \\
			\midrule
                CoCoOp~\citep{cocoop} &{80.47} & 71.69 &75.83 \\
                \midrule
                w/ MLP &{79.18} & 71.98 &75.41 \\
                w/ Transformer &{80.17} & 72.03 &75.88 \\
                \midrule
               w/ GAN &{81.15} & 71.44 & 75.99 \\
                w/ VAE &{80.73} & 72.09 &76.17 \\
                w/ Normalizing flows &{80.65} & 72.43 & 76.32 \\
            \rowcolor{lightGray}
			{\textbf{w/ Diffusion} }   & \color{blue}{81.35}  & \color{blue}{74.97}  &\color{blue}{78.02}   \\   
			\bottomrule
		\end{tabular}}
  \vspace{-4mm}
\label{tab:generative}
\end{wraptable} 
prompts $\bm{V}^*$, are used to supervise and train these three types of generative models.
Table~\ref{tab:generative} shows  the diffusion model 
outperforms all variants in accuracy. Specifically, our proposed per-sample overfitting integration with MLP and transformer architectures shows a slight harmonic mean improvement over the CoCoOp
baseline, validating the effectiveness of our per-sample prompt overfitting. 
Notably, our diffusion model presents a good increase in accuracy, surpassing the GAN, VAE and normalizing flows models by 2.01\%,  1.85\% and 1.70\%, respectively.


%
\begin{figure}[t]
    \centering
    \begin{minipage}{0.45\textwidth}
        \centering
        \vspace{-5mm}
        \includegraphics[width=1.\textwidth]{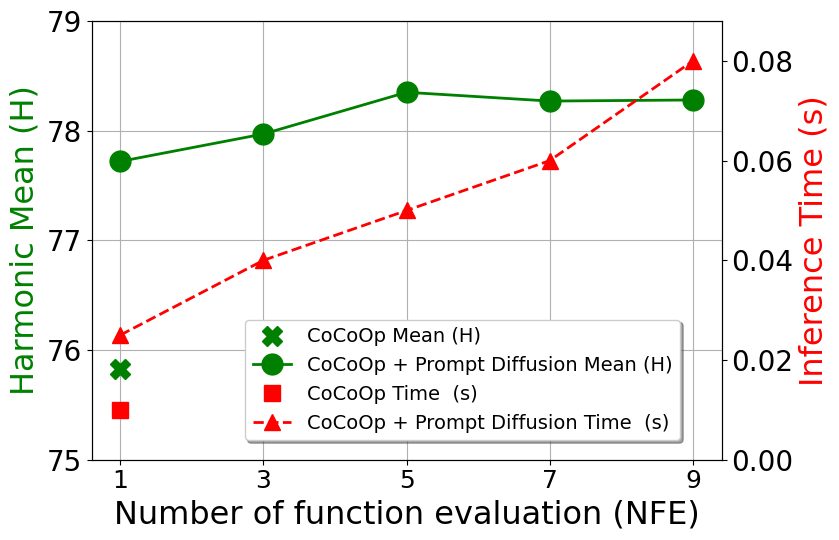}
        \caption{\textbf{Effect of number of function evaluation} on base-to-new generalization.}
                \vspace{-5mm}
        \label{fig:uncertainty}
    \end{minipage}%
    \hfill
    \begin{minipage}{0.45\textwidth}
        \centering
        \includegraphics[width=.95\textwidth]{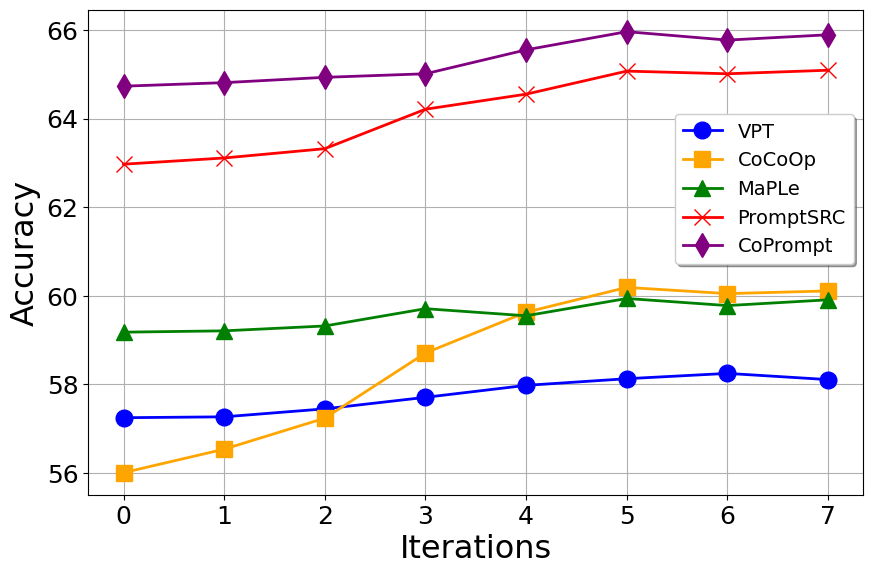}

\caption{\textbf{Impact of iterations} on per-sample prompt overfitting for novel classes.}
        \label{fig:iteration_overfitting}
    \end{minipage}
\end{figure}

\begin{figure}[t]
                \centering
            \includegraphics[width=1.\linewidth]{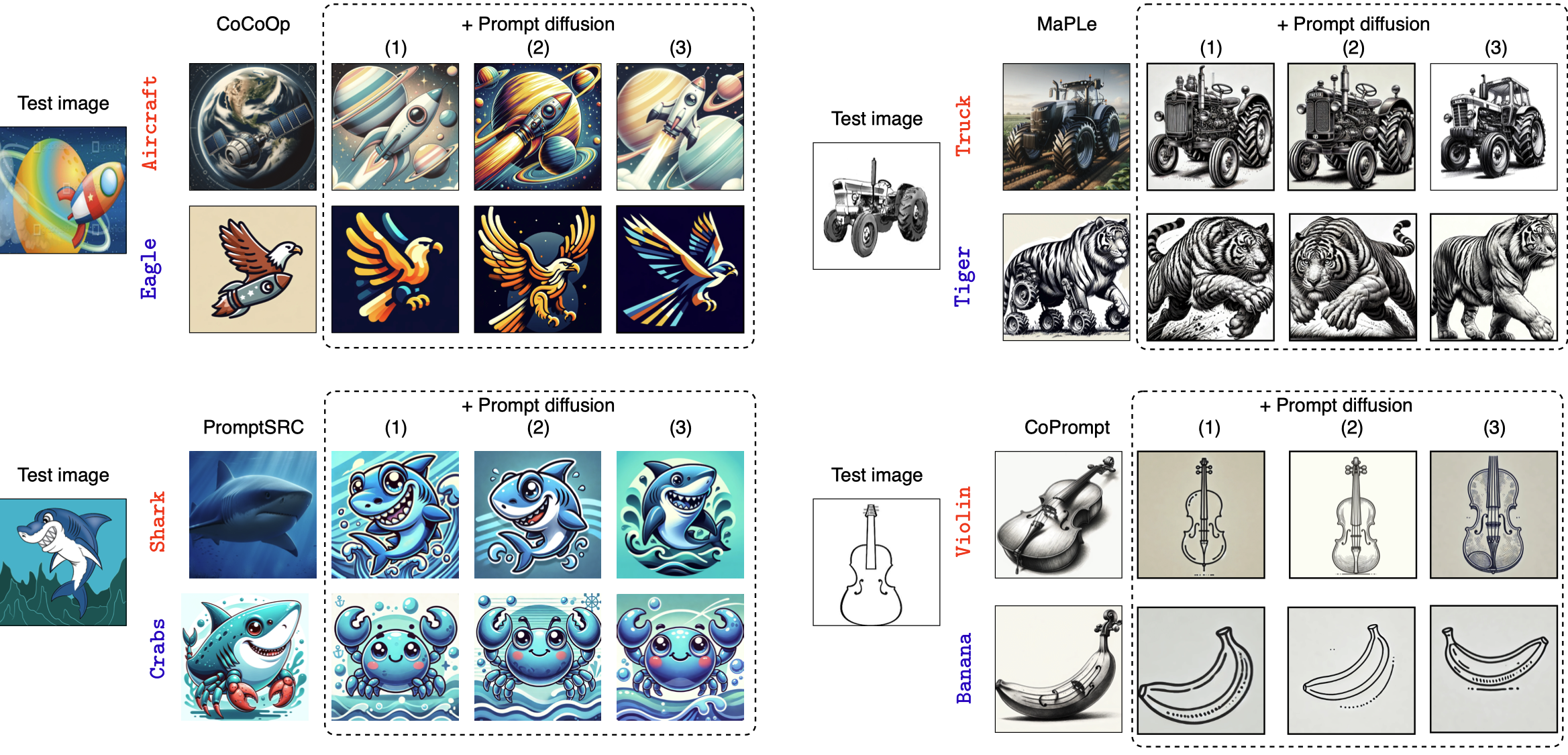}
     \caption{\textbf{Visualization of generated prompts} by ControlNet~\citep{zhang2023adding}.
        We generate diverse images by utilizing three distinct Monte Carlo prompt samples, each derived from our prompt distribution and based on varying random noise. When using ground truth names (\textcolor{red}{\texttt{red}} names), images produced by CoCoOp, MaPLe, PromptSRC, and CoPrompt are more realistic, while prompt diffusion incorporates domain-specific details from the test image. Regarding the other classes (\textcolor{blue}{\texttt{blue}} names), CoCoOp, MaPLe, PromptSRC, and CoPrompt blend true class features with others, potentially leading to confusion, but our plugin using these methods can generate a stylized version of the specified class.
        This suggests that our plugin enables the distilling of unique domain details from the test image without conflating them with class labels.}
                \vspace{-7mm}
        \label{fig:stable}
\end{figure}

\noindent\textbf{Effect of the number of function evaluation.}
Our prompt diffusion utilizes the fast ODE-based
sampling strategy introduced by~\citep{zhou2024fast}
enabling efficient sampling with a reduced number of timesteps during testing. In Figure~\ref{fig:uncertainty}, we analyze the effect of different numbers of function evaluations (NFE) on both final performance and inference time. Our findings indicate that at an NFE of 5, our method achieves the best balance between performance and prediction time. In comparison to the original CoCoOp, our approach results in only a 0.045-second increase in prediction time while delivering a substantial performance improvement. This highlights the effectiveness of our method in balancing accuracy and computational efficiency. 

\noindent\textbf{Impact of iterations on per-sample prompt overfitting.} In our prompt diffusion method, per-sample prompt overfitting is crucial to generate optimal prompts during training. Figure~\ref{fig:iteration_overfitting} shows that as iterations increase, accuracy on novel classes improves for all methods, peaking at iteration 5. This shows that the quality of the optimal prompt directly influences the final performance. Moreover, our prompt diffusion effectively learns the transformation from a vanilla prompt to an optimal prompt throughout training using a diffusion transformer. As a result, during testing, our method can generate a sample-specific prompt for each test sample, thereby improving accuracy.

\textbf{Visualization of generated prompts.} We also visualize the generated per-sample prompts during inference in Figure~\ref{fig:stable}, demonstrating  our diffusion prompting method effectively distills unique domain details from the test image without mixing them with class labels. 
This shows the better capability of the diffusion model in refining the prompt learning process for vision-language tasks.
\section{Conclusion}
Our approach addresses the limitations of fixed prompts by introducing a method that crafts customized prompts for individual test samples, enhancing model robustness against distributional shifts. The diffusion model serves as the backbone of this method, enabling a generative process that refines prompts from a random initialization to an optimized state, tailored to each specific instance. The versatility and modality-agnostic nature of prompt diffusion mark it as an universally applicable solution that integrates smoothly with existing prompt learning methods, regardless of the data type. The empirical results across a wide range of datasets validate the efficacy of our method, demonstrating its increased robustness in generalization tasks.


\bibliography{ref}
\bibliographystyle{iclr2025_conference}

\clearpage
\appendix

\begin{figure}[t]
    \centering
    \includegraphics[width=1.\linewidth]{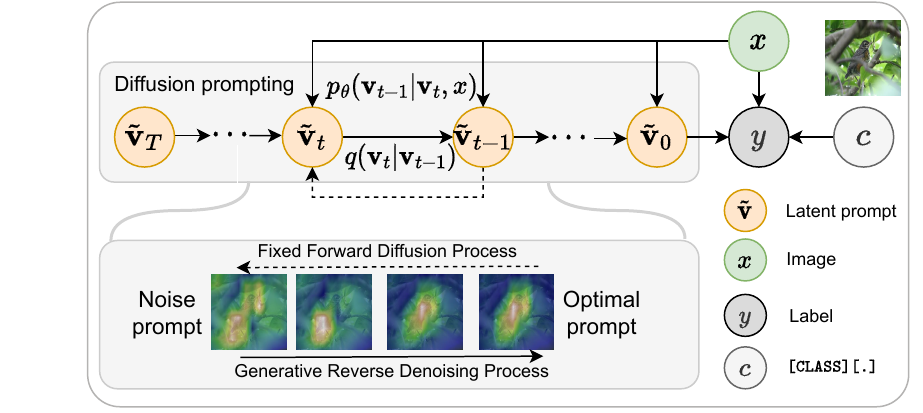}
    \caption{\textbf{Computational Graph and Diffused Prompt}. The top diagram illustrates the computational structure of our method. On the bottom left, we showcase the graphical representation of our method's diffused prompt. Through diffusion sampling techniques, the diffused prompt $\mathbf{V}_{t-1}$ emerges as a fusion of $\mathbf{V}_{t}$ and $x$. The resultant prediction $y$ is then formed by leveraging the diffused prompt, expanded as "\texttt{[CLASS]}" label $\mathbf{c}$: $\{\mathbf{V}_{0}, \mathbf{c}\}$, and paired with the image descriptor $x$. Within the shaded rectangle, dashed arrows denote the diffusion procedure, while solid arrows highlight the sampling steps.
}
    \label{fig:graph}
\end{figure}

\section{Computational graph of diffusion prompting}
In this section, we illustrate the computational graph of diffusion prompting in Figure~\ref{fig:graph}. The figure is divided into two parts: the top diagram displays the overall computational structure of the method, while the bottom left part presents a graphical representation of the method's diffused prompt. In this process, the diffused prompt \(\mathbf{V}_{t-1}\) is created through a fusion of \(\mathbf{V}_{t}\) and the image descriptor \(x\) using diffusion sampling techniques. This results in the prediction \(y\), which is generated by combining the diffused prompt, expanded as the "\texttt{[CLASS]}" label \(\mathbf{c}\): \(\{\mathbf{V}_{0}, \mathbf{c}\}\), with the image descriptor \(x\). The shaded rectangle in the diagram helps to visually differentiate the components of the process, where dashed arrows indicate the diffusion steps, and solid arrows represent the sampling stages. This figure provides a clear and concise visual representation of the complex processes involved in diffused prompting, highlighting the intricate interactions between different components of the computational model.

\section{Additional related works}
\noindent\textbf{Diffusion models.} This class of neural generative models is characterized by the employment of stochastic diffusion processes akin to those observed in thermodynamic systems \citep{sohl2015deep,song2020denoising}. The operational principle of these models involves a sequential noise addition to data samples, followed by a learned neural network's effort to reverse this process. This is achieved by gradually denoising the noise-saturated sample to retrieve data reflecting the trained data distribution. Significant strides in the realm of image generation have been accredited to the works of Ho \etal~\citep{ho2020denoising} and Song \etal~\citep{song2020denoising}, while Dhariwal and Nichol \citep{dhariwal2021diffusion} have been pivotal in pioneering classifier-guided diffusion for generation under specific conditions. Building on this foundation, GLIDE \citep{nichol2021glide} has further refined the methodology by incorporating conditioning on textual representations derived from CLIP. The concept of classifier-free guidance introduced by Ho \etal~\citep{ho2022classifier} has brought forward a method of conditioning that judiciously balances fidelity and diversity, leading to notable enhancements in model performance \citep{nichol2021glide}. However, guided diffusion models typically necessitate an extensive corpus of image-annotation pairs for effective training, prompting Hu \etal~\citep{hu2022self} to suggest the novel concept of self-guided diffusion models. More contemporary developments include Hyperdiffusion \citep{lutati2022ocd, erkocc2023hyperdiffusion}, which targets the generation of implicit neural representations and 3D reconstruction through diffusion in weight space.
To the best of our knowledge, we are the first to introduce diffusion models into the realm of prompt learning.  Our diffusion prompting involves gradually refining prompts with a diffusion transformer, which leads to the development of custom prompts tailored to each sample, thereby enhancing the accuracy of predictions and their generalization across downstream tasks. 

\section{Hyperparameter sensitivity and few-shot.}
Table~\ref{tab:epoch} presents a comparison of our model's performance over different epochs relative to CoOp's training duration of 200 epochs. Our model reaches convergence around the 50-epoch mark and surpasses the performance of CoOp after 200 epochs. Additionally, we also conduct a few-shot learning experiment (4-shot) similar to those conducted with CoOp and CoCoOp, as shown in Table~\ref{tab:dg_4shot}. In these comparisons, our model consistently achieves improved performance across a range of datasets.

\section{Parameter-efficient comparison.} 
Table~\ref{tab:pfft} contrasts our approach with four other parameter-efficient fine-tuning techniques. Our integration with MaPLe~\citep{khattak2023maple} showcases superior average performance, underscoring its superior ability to generalize in comparison to other parameter-efficient fine-tuning approaches. Furthermore, we have applied our plugin in conjunction with LLU~\citep{ibing2023localized}  in a base-to-new setting, where it also exhibits enhanced performance relative to LLU alone.

\begin{table}[t]
\centering
    \caption{Comparison with CoOp for various epochs.}
\scalebox{1.}{
		\begin{tabular}{lcc|c}
			\toprule
			Epochs & Base & New & H \\
			\midrule
                10 & 80.29 & 73.51 & 76.26\\
                50 & 81.35 & 74.97 & 77.72\\
                100 & {81.47} & 74.88 & 77.70\\
			200 & {81.78}  & 74.24  & 77.65  \\   
			\bottomrule
		\end{tabular}}
    \label{tab:epoch}
\end{table}

\begin{table}[t]
    \centering
        \caption{Comparison with 4-shots on domain generalization. Our results are competitive for all domains. 
        }
    \scalebox{1.}{
	\begin{tabular}{l cccccc}
		\toprule
		& Source & \multicolumn{5}{c}{Target} \\ \cmidrule(lr){2-2} \cmidrule(lr){3-7}
		& ImageNet & -V2 & -S & -A & -R &Average\\

  		\midrule
  VPT~\citep{jia2022visual}  & 69.24 & 62.13 & 45.78 &48.16 &72.91 &57.37\\
		            \rowcolor{lightGray} {{+ Prompt Diffusion}} &{70.27}        &{63.83}           &{48.15}            &{50.97}   &{76.15} &{60.12} \\
		\midrule
  CoCoOp~\citep{cocoop}  & 70.13 & 63.05 & 46.48 &49.36 &73.80 &58.17\\
		            \rowcolor{lightGray} {{+ Prompt Diffusion}} & {70.96}        & {64.12}           & {48.92}            &{51.47}   & {76.93} &{60.75} \\
		\midrule
		{MaPLe}~\citep{khattak2023maple} & 70.72 & {64.07} &{49.15} & {50.90}  & {76.98}  & {60.83}  \\
  		            \rowcolor{lightGray} {{+ Prompt Diffusion}}  & \color{blue}{71.23}        &\color{blue}{65.24}           &\color{blue}{50.21}            &\color{blue}{51.93}  &\color{blue}{78.06} &\color{blue}{62.11} \\
                
		\bottomrule 
	\end{tabular}}
\label{tab:dg_4shot}
\end{table}

\begin{table}[t]
\centering
 \caption{Comparison with parameter-efficient fine-tuning methods in the base-to-new setting  across 11 datasets.}
\scalebox{1.}{
		\begin{tabular}{l|c | cc|c}
			\toprule
			& Venues & Base & New & H \\
			\midrule
                ProGrad~\citep{zhu2023prompt} & ICCV 23 & 82.79 & 68.55 & 74.46\\
                CLIP Adapter~\citep{gao2024clip} & IJCV & 82.62 & 70.97 & 76.02\\
                LLU~\citep{ibing2023localized}   & CVPR 23 & 83.48 & 74.47 & 78.46\\
                MaPLe~\citep{khattak2023maple}  & CVPR 23 & 82.28 & 75.14 & 78.55\\ \midrule
                      \rowcolor{lightGray}
                      			{{{LLU + Diffusion Prompt}}}  &  &\color{blue}{84.45}   &  75.99  & 79.17 \\   
                         \rowcolor{lightGray}
			{{{MaPLe +  Diffusion Prompt}}}  &  &{83.39}  &\color{blue}{77.12}  &\color{blue}{79.96}  \\         
			\bottomrule
		\end{tabular}}
    \label{tab:pfft}
\end{table}

\end{document}